\documentclass[11pt]{article}
\usepackage[final]{acl}
\usepackage{times}
\usepackage{latexsym}
\usepackage[T1]{fontenc}
\usepackage[utf8]{inputenc}
\usepackage{microtype}
\usepackage{inconsolata}
\usepackage{amsmath,amssymb}
\usepackage{booktabs}
\usepackage{multirow}
\usepackage{makecell}
\usepackage{graphicx}
\usepackage{xcolor}

\newcommand{\thd}[1]{\textbf{#1}}
\newcommand{\method}{\textsc{NestOne}}   
\newcommand{\intern}{Intern-Decision}

\newcommand{\KL}{\mathrm{KL}}

\title{From Large Language Models to Efficient System One Models via Adaptive Computation}

\author{
  Minh-Anh Nguyen$^{1}$\\
  $^{1}$Nanyang Technological University, Singapore
}

\begin{document}
\maketitle
\begingroup
\renewcommand\thefootnote{}\footnotetext{Preprint. Work in progress.}
\addtocounter{footnote}{-1}
\endgroup

\begin{abstract}
Transferring LLMs into System One decision models typically replaces the language modeling head with a decision head while retaining the full backbone, making every decision pay for all layers. Yet it remains unclear how much depth such decisions require. We find that decision information is concentrated in intermediate layers, with early layers building essential features and the final six layers removable without loss, while independently trained smaller models largely mimic the full model rather than developing specialized decision capabilities. These findings suggest that efficiency requires not simply smaller backbones, but shared computation across different depths. Building on this, we propose \method{}, which jointly trains a full model with prefix sub-models that share its final layers and decision head through self-distillation. A lightweight gate reads the hidden state at each exit to determine whether additional computation is needed, allowing one model to operate at multiple layers. Across seven decision benchmarks, our approach matches or outperforms independently trained models at every depth, while achieving better accuracy-efficiency trade-offs through adaptive computation.
\end{abstract}

\begin{figure}[t]
\centering
\includegraphics[width=\linewidth]{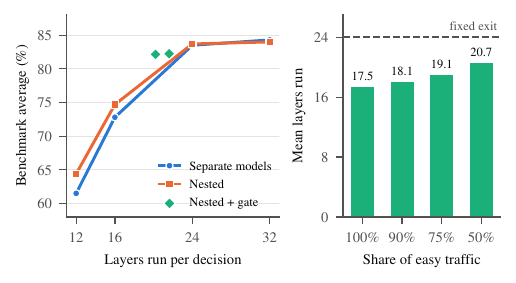}
\caption{Left: Accuracy versus computation depth for independently trained models, and \method{} w/o the stop gate. Right: Average depth of the gate versus the share of easy decisions; on the four largest benchmarks, accuracy stays within 0.1 points of the fixed 24-layer exit. All models use Qwen3.5-4B and the same training data; models are averaged over 3 seeds.}
\label{fig:teaser}
\end{figure}

\section{Introduction}
\label{sec:intro}

Dual-process theories distinguish fast, intuitive \emph{System One} thinking from slow, deliberate \emph{System Two} thinking~\citep{kahneman2011thinking}. LLM-based agents increasingly require both. Alongside long chain-of-thought reasoning, they make frequent, lightweight decisions, such as routing requests, selecting tools, flagging unsafe prompts, and recognizing intents, often under tight latency constraints~\citep{typesafe2026jev}. System One decision models address this need by fine-tuning an LLM to predict a distribution over candidate options at a single position, without generating tokens~\citep{intern_decision,jeva}. For example, \intern{}-4B, built on Qwen3.5-4B~\citep{qwen35}, achieves 90.0\% macro accuracy across seven decision benchmarks. However, replacing the language modeling head with a decision head leaves the full backbone intact: every decision still traverses all 32 layers, incurring a latency of approximately 1.45\,ms per layer at batch size one in our setting.

This raises a fundamental question: \emph{how much computational depth does a System One decision actually require?} Fine-tuning is necessary to equip an LLM with decision capabilities that its base model lacks (Section~\ref{sec:observation}), but whether all inherited layers are necessary remains unclear. A natural way to reduce inference cost is to truncate the backbone, trading computational depth for potentially lower accuracy. As shown in the left panel of Figure~\ref{fig:teaser}, this trade-off is highly non-uniform: reducing the model from 32 to 24 layers largely preserves performance, whereas further truncation causes substantial degradation, with accuracy dropping from 84.3\% to 72.8\% and 61.5\% at 16 and 12 layers, respectively. This highlights the limitation of fixed-depth inference, which imposes the same computational budget on all inputs. Yet, not every decision requires the same amount of computation. This motivates adaptive depth allocation (Figure~\ref{fig:teaser}, right), which allows computation to stop once sufficient depth is reached, reducing inference cost while preserving accuracy.

We find that decision computation is concentrated in intermediate layers, while independently trained smaller models largely replicate the full model rather than develop specialized decision capabilities (Section~\ref{sec:observation}). First, decisions emerge within a narrow band of middle layers: integrated-gradient attribution places 50\% of the decision by layer 20 and 80\% by layer 23. Consistently, ablating layers 24--29 leaves accuracy unchanged, while ablating layers 15--23 eliminates the decision margin (Section~\ref{subsec:obs_where}). Second, independently trained smaller models behave more like truncated full models than independent specialists: 89--92\% of their correct decisions are already made by the full model, and their distinct wins exceed full-model seed disagreement by only 2--3 points, mostly on low-confidence decisions (Section~\ref{subsec:obs_subset}). These findings show that training smaller models or truncating the backbone alone is insufficient. Instead, decision models should learn to make reliable decisions at multiple depths, adapting computation to input difficulty.

Based on these findings, we propose \method{}, a framework that transforms an LLM into a multi-depth System One decision model. \method{} jointly trains the full model and nested prefix sub-models with shared final layers and decision head, combining decision supervision and self-distillation from the full model. This enables a single training run to produce multiple exits within one checkpoint. At each exit, a lightweight gate reads the hidden state to determine whether further computation is needed before producing the decision. Our contributions are threefold: (i) We show that decision information is concentrated in intermediate layers and that independently trained smaller models largely reproduce, rather than specialize in, the full model's decisions; (ii) We propose \method{}, a nested multi-depth training framework that combines decision supervision, self-distillation, and a hidden-state stopping gate for transfering from LLMs to System One Models efficiency; (iii) We demonstrate that \method{} matches or outperforms independently trained models at every depth while reducing inference cost through adaptive early exiting, using a single training run to obtain multiple models.

\section{Background and Related Work}
\label{sec:background}

\paragraph{System One Decision Models.} System One models take a state, a typed question, and a set of options, and return a distribution over the options in a single forward pass without generating text~\citep{typesafe2026jev}. Following Jev, a growing number of such models are built by adapting a pretrained LLM. Open recipes read the option-symbol logits of an instruction-tuned LLM at the last position, with no training, or add LoRA adapters and a decision head initialized from the language-model head~\citep{geuter2026inner}. \intern{}~\citep{intern_decision} fully fine-tunes Qwen3.5~\citep{qwen35} at 0.8B, 2B, and 4B parameters, Microsoft-Decision-1 is post-trained from Qwen3.5-9B~\citep{microsoft2026decision}, CLM trains a decision model contrastively from Qwen3-8B~\citep{kwok2026clm}, and CS1 trains a 4B model on 12 million clinical decisions~\citep{zhang2026cs1}. Other work changes how options are read out. Jeva encodes the request and each option separately so that the decision is permutation-equivariant, can abstain, and can reuse option states across requests~\citep{jeva}; Prism-1 scores each option in its own attention block so that neither option order nor added rivals change the result~\citep{teng2026prism}; and Consider-then-Choose removes order dependence at inference without training~\citep{huang2026consider}.

\paragraph{Evaluating System One Models.} A second line of work asks where System One models can replace LLMs. System1-Arena compares ten System One models with open and closed LLMs through a shared decision interface and finds that the fast models suffice for many single-step decisions but fall behind on long-horizon tasks~\citep{chen2026system1arena}. Other studies evaluate System One models as decision points in agent harnesses~\citep{li2026fastmodels}, and test their transfer to low-resource languages~\citep{fadli2026transfer}. These studies treat each model as a black box and measure what it outputs.

\paragraph{Layer Pruning.} The deeper layers of LLMs are known to be partly redundant. Removing a block of late layers costs little on question answering~\citep{gromov2024unreasonable,men2024shortgpt}, and merging or pruning layers followed by distillation recovers most of the original accuracy at a smaller size~\citep{yang2024laco,muralidharan2024compact,xia2024sheared}. These methods produce one smaller model with one fixed depth, and they select layers by similarity or importance scores computed for generation rather than by where a decision is computed.

Existing work on System One models focuses on how to train an LLM into a decision model, whether through readout design, training data, or architecture, and on how the resulting models behave. Each of these models keeps the full depth of its backbone, and cost is reduced only by choosing a smaller backbone. To our knowledge, no prior work asks where inside the LLM the decision is computed, or what a smaller decision model learns relative to the full one. We address both questions first (Section~\ref{sec:observation}). The answers then guide an efficient way to turn one LLM into a System One model with several depths: nested prefix training with self-distillation, and a gate that reads the hidden state before the readout is computed (Section~\ref{sec:method}).

\section{Understanding Depth in System One Models}
\label{sec:observation}

We investigate what enables an LLM to become a System One decision model by comparing the model before and after decision fine-tuning, along with smaller variants obtained by progressively removing layers from the fine-tuned model. Specifically, we ask two questions: (i) Where in the network are decisions computed (Section~\ref{subsec:obs_where})? (ii) What do smaller models learn compared to the full model (Section~\ref{subsec:obs_subset})?

\textbf{Setting.} We study Qwen3.5-4B~\citep{qwen35}, whose 32 layers we index 0--31, and \intern{}-4B~\citep{intern_decision}, a released decision model built on the same backbone. To establish a controlled reference, we re-train Qwen3.5-4B using the same training recipe as \intern{}-4B. We refer to this re-trained model as \textbf{full32}, which serves as the starting point and reference for all our models. Specifically, full32 is fully fine-tuned with cross-entropy on the option symbols (AdamW, learning rate $2\times10^{-6}$ with cosine decay, 2 epochs, about 131K tokens per step) on 44,941 decisions drawn from public training splits~\citep{zhang2015character,liu2024toolace,lin2023toxicchat,ji2023beavertails,casanueva2020efficient,fitzgerald2023massive,clark2019boolq} and deduplicated against all test states. It reaches $84.3\pm0.2\%$ over three seeds, compared with 90.0\% for \intern{}-4B, whose larger training set covers the J-Hard and Typed benchmarks better. We consider two layer-cutting strategies: a middle cut, $\mathrm{M}k$, which retains layers $0,\ldots,k{-}3$ and the final two layers (30--31), and a tail cut, $\mathrm{T}k$, which retains only the first $k$ layers ($0,\ldots,k{-}1$).

\begin{figure*}[t]
\centering
\includegraphics[width=\linewidth]{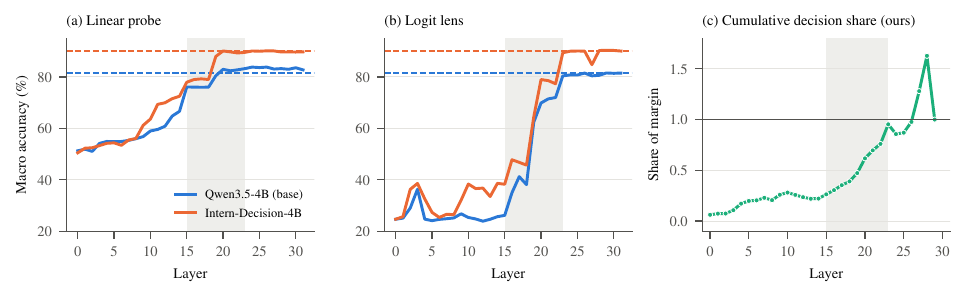}
\caption{Where the decision becomes readable and where it is written. (a) Linear-probe accuracy for the gold option at the readout position of each layer, and (b) logit-lens accuracy using each model's own head, macro-averaged over the seven benchmarks; dashed lines mark each model's own accuracy. (c) Cumulative share of the decision margin contributed by layers up to $\ell$ in full32, measured with integrated gradients. The shaded band marks layers 15--23.}
\label{fig:where}
\end{figure*}

\subsection{The Decision Is Written at Mid Depth}
\label{subsec:obs_where}

\paragraph{When Does the Decision Become Readable?}
We examine when the decision-related information emerges among layers and when it becomes accessible to the model's own readout. At each layer, we extract the residual state at the readout position and evaluate it using two complementary methods. The \emph{logit lens}~\citep{nostalgebraist2020logitlens} applies the model's final normalization and output head to the intermediate state, measuring whether the model can already produce the correct answer at that depth. In contrast, a \emph{linear probe}~\citep{alain2017understanding}, trained to predict the gold option from the same state, measures how much decision-relevant information is linearly recoverable, regardless of whether the model's own readout can access it. Thus, the gap between the probe and logit-lens accuracies indicates information that has been encoded in the residual stream but is not yet usable by the model's output head.

Figure~\ref{fig:where}(a,b) shows a consistent pattern across the base model and \intern{}-4B. At layer 15, the probe substantially outperforms the logit lens (76.1\% vs.\ 26.2\% for the base model; 77.9\% vs.\ 38.3\% for \intern{}-4B), indicating that decision information is already encoded but not yet accessible to the model's readout. This gap closes over layers 19--23: for \intern{}-4B, the probe reaches 90.1\% at layer 20, matching its final accuracy (90.0\%), while the logit lens rises from 64.4\% at layer 19 to 89.4\% at layer 23. The two readouts then differ by only 0.1 points (vs.\ 2.8 points for the base model). Neither improves beyond layer 23, suggesting that layers 24--31 add no linearly recoverable decision information. Fine-tuning raises the probe plateau by about seven points (89.5\% vs.\ 83.2\% at layer 23) without shifting the depth at which decisions become readable.

\paragraph{Which Layers Shape the Decision?} Readability shows where the answer is available, not which layers build it there. We therefore decompose the readout state, i.e.\ the residual stream at the readout position after the last body layer, as $\mathbf{h}=\mathbf{h}_0+\sum_{\ell=0}^{29}\Delta_\ell$, where $\mathbf{h}_0$ is the token embedding and $\Delta_\ell$ is the \emph{contribution} of layer $\ell$, i.e.\ the sum of the attention and MLP outputs that it adds to the residual stream. Let $s_o(\mathbf{h})$ be the logit of option $o$ obtained by passing $\mathbf{h}$ through decision head. With gold option $y$ among $K$ options, the decision margin is
\begin{equation*}
m(\mathbf{h})=s_y(\mathbf{h})-\frac{1}{K-1}\sum_{o\neq y}s_o(\mathbf{h}),
\end{equation*}
so a larger margin puts more probability on the gold option. Integrated gradients (IG)~\citep{sundararajan2017axiomatic} attribute the margin to these contributions. Along the straight path $\mathbf{h}(a)=\mathbf{h}_0+a(\mathbf{h}-\mathbf{h}_0)$ from the embedding to the readout state, layer $\ell$ receives
\begin{equation*}
\mathrm{IG}_\ell=\Delta_\ell^\top\int_0^1 \nabla_{\mathbf{h}} m\bigl(\mathbf{h}(a)\bigr)\,da,
\end{equation*}
its contribution multiplied by the average sensitivity of the margin along the path. A positive $\mathrm{IG}_\ell$ means that layer $\ell$ moves the decision distribution toward the gold option, and a negative one moves it away. The attributions add up to the total margin change, $\sum_\ell\mathrm{IG}_\ell=m(\mathbf{h})-m(\mathbf{h}_0)$. 

Figure~\ref{fig:where}(c) reveals three distinct regimes in full32. Early layers (0--14) account for only 22\% of the final margin, with no individual layer contributing more than 7\%. Middle layers (15--23) contribute another 73\%, raising the cumulative margin from 22\% to 95\%. In contrast, late layers exhibit substantial but largely canceling updates: layers 27 and 28 contribute 30\% and 35\%, respectively, raising the running margin to 1.6$\times$ its final value, before layer 29 reverses this increase ($-63\%$). Consequently, layers 24--29 contribute only 5\% net, primarily rescaling the margin without changing the preferred option. Overall, the middle layers play the dominant role in shaping the final decision distribution of a fine-tuned System One model.

\begin{table}[t]
\centering
\small
\setlength{\tabcolsep}{4pt}
\begin{tabular}{@{}llrr@{}}
\toprule
\multicolumn{4}{@{}l}{\textit{(a) Removing layers (\intern{}-4B, 90.05)}}\\
Cut & Removed & Acc & $\Delta$ [95\% CI] \\
\midrule
M26 & 24--29 & 90.1 & $+0.0$ [$-0.9$, $+0.9$] \\
T24 & 24--31 & 89.4 & $-0.6$ [$-1.2$, $-0.1$] \\
M24 & 22--29 & 88.8 & $-1.2$ [$-2.1$, $-0.4$] \\
T22 & 22--31 & 78.5 & $-11.6$ [$-14.1$, $-9.3$] \\
\midrule
\multicolumn{4}{@{}l}{\textit{(b) Mean ablation (full32)}}\\
Ablated & & Margin & $\Delta$Acc \\
\midrule
0 & & $95\%$ & $-46.6$ \\
6 & & $31\%$ & $-7.8$ \\
\midrule
24--29 & & $-6\%$ & $+1.7$ \\
20--23 & & $34\%$ & $-2.3$ \\
19 + 23 & & $53\%$ & $-5.5$ \\
15--23 & & $102\%$ & $-57.8$ \\
\bottomrule
\end{tabular}
\caption{(a) Accuracy after removing layers without retraining. (b) Share of the mean decision margin removed, and accuracy change, when the contribution of a single early layer (0, 6) or jointly of a group of layers is replaced by its mean (348 decisions, 50 per benchmark).}
\label{tab:remove}
\end{table}

\paragraph{Which Layers Can Be Removed?}
Table~\ref{tab:remove} evaluates layer importance through two complementary interventions. First, we remove layers without retraining (Table~\ref{tab:remove}a), using a middle cut $\mathrm{M}k$ or a tail cut $\mathrm{T}k$. Second, we perform \emph{joint mean ablation} (Table~\ref{tab:remove}b): for each layer $\ell$, we estimate its mean contribution $\bar{\Delta}_\ell$ across token positions and approximately 1,700 decisions (up to 400 per benchmark), then replace each layer's contribution with this mean at every position. This preserves the residual stream's typical scale while removing decision-specific information. Joint ablation measures the group's collective contribution to decision-making, accounting for interactions among its layers while preventing them from compensating for one another. We report the resulting change in mean margin and accuracy over 348 decisions.

The two interventions reveal distinct layer roles. In \intern{}-4B, removing layers 24--29 without retraining leaves accuracy unchanged, while removing layers 24--31 costs only 0.6 points. Similarly, full32 retains 84.4\% accuracy under $\mathrm{T}24$, compared with 84.2\% uncut; cutting into the middle band is substantially more damaging, with $\mathrm{T}22$ losing 11.6 points. Mean ablation confirms this pattern: jointly ablating layers 24--29 slightly improves margin and accuracy, whereas ablating layers 15--23 eliminates the entire margin and reduces accuracy by 57.8 points. Early layers, however, remain essential despite their small direct margin contributions: ablating layer 0 costs 46.6 points and layer 6 costs 7.8, reflecting their role in constructing features used by later layers. Consistent with this, a 16-layer model trained only on layers 8--21 and 30--31, the layers most affected by fine-tuning, achieves just 35.7\%, compared with 73.8\% for the $\mathrm{M}16$ prefix model.

\textbf{Takeaway.} A decision-tuned LLM writes its decision in a band of middle layers (15--23) on top of feature-building early layers, and its last body layers carry no decision. 

\begin{table}[t]
\centering
\small
\setlength{\tabcolsep}{3pt}
\begin{tabular}{@{}lrrrrrrr@{}}
\toprule
 & & \multicolumn{2}{c}{Only correct} & & & \multicolumn{2}{c}{Acc $\mid p_{\text{full}}$} \\
\cmidrule(lr){3-4}\cmidrule(l){7-8}
Model & Acc & full & small & Sub. & Rec. & $\geq$.8 & $<$.2 \\
\midrule
\multicolumn{8}{@{}l}{\textit{Noise floor: other seeds of full32}}\\
seed 43 & 84.1 & 1.5 & 1.4 & .977 & .977 & 100.0 & 0.7 \\
seed 44 & 84.5 & 1.8 & 2.0 & .968 & .972 & 100.0 & 0.9 \\
\midrule
T24 cut & 84.5 & 1.5 & 1.7 & .975 & .975 & 99.9 & 1.7 \\
M24 & 83.5 & 2.7 & 1.9 & .967 & .959 & 99.7 & 2.0 \\
M16 & 73.8 & 14.4 & 3.8 & .914 & .795 & 84.1 & 16.7 \\
M12 & 62.9 & 25.6 & 4.3 & .891 & .666 & 70.4 & 19.5 \\
\bottomrule
\end{tabular}
\caption{Decision-level comparison with full32 (seed 42). \thd{Noise floor}: full32 trained with seeds 43 and 44, compared in the same way. \thd{Only correct}: share of decisions answered correctly by only full32 or the smaller model. \thd{Sub.} and \thd{Rec.}: $P(\text{full correct}\mid\text{small correct})$ and $P(\text{small correct}\mid\text{full correct})$, respectively. \thd{Acc $\mid p_{\text{full}}$}: smaller-model accuracy when full32 assigns the gold option probability $\geq0.8$ or $<0.2$. M24, M16, and M12 are independently trained; T24 is truncated from full32 without retraining. All rates are macro-averaged across benchmarks.}
\label{tab:overlap}
\end{table}

\begin{figure*}[t]
\centering
\includegraphics[width=\linewidth]{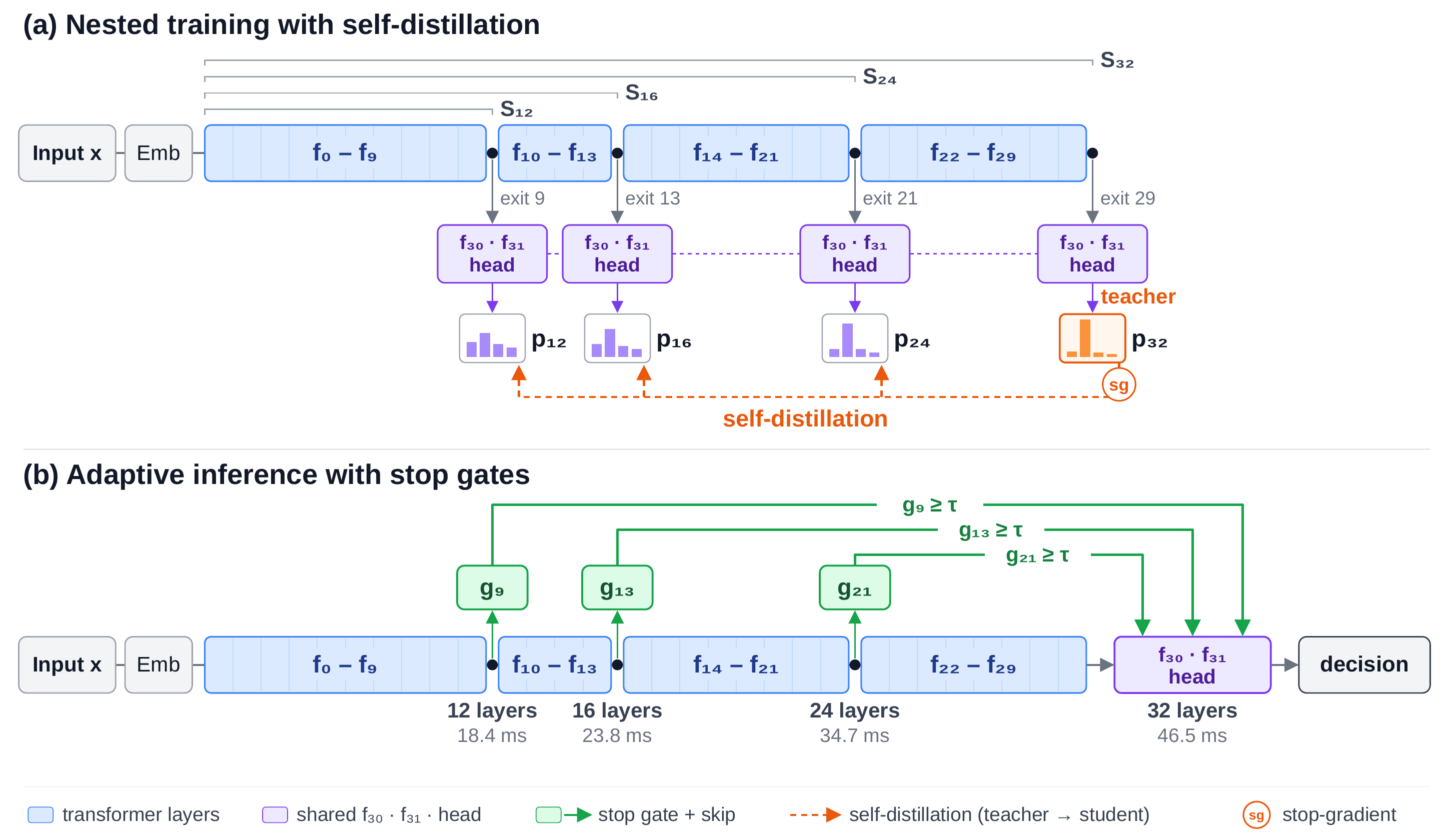}
\caption{\textbf{Overview of \method{}.}
 \textbf{(a) Nested training with self-distillation.} Prefix exits share the full model's weights and final block. The full model learns from the gold option $y$, while each sub-model learns from both $y$ and the full model's distribution. \textbf{(b) Adaptive inference with stop gates.} The model runs layer by layer, with a lightweight gate evaluating the hidden state at each intermediate exit. If $g_e\geq\tau$, it runs the final block and returns a decision; otherwise, it continues from the saved states without recomputation.}
\label{fig:method}
\end{figure*}

\subsection{Diagnosing Decision Behavior}
\label{subsec:obs_subset}

We next examine what independently trained smaller models learn relative to full32. We fine-tune M24, M16, and M12 using the same data and recipe as full32, obtaining 83.5\%, 73.8\%, and 62.9\% accuracy, respectively. We compare their predictions with full32 decision by decision, categorizing each as both correct, only full32 correct, only the smaller model correct, or both incorrect. To establish a noise floor, we also compare full32 against two independently trained seeds (43 and 44), which uniquely solve 1.4--2.0\% of decisions missed by the reference seed (Table~\ref{tab:overlap}, top).

\paragraph{Smaller Models Mostly Reproduce Full32's Decisions.}
Table~\ref{tab:overlap} shows that M24 rarely makes a correct prediction that full32 misses: 1.9\% of decisions, within the noise floor of 1.4--2.0\%, and its subset score (.967) matches that of another full32 seed (.968--.977). M16 and M12 remain largely subsets of full32: 91\% and 89\% of their correct predictions, respectively, are also correct under full32. Moreover, they lose substantially more decisions than they uniquely solve, by roughly a factor of three to four (1,012 vs.\ 352 for M16; 1,513 vs.\ 425 for M12).

\paragraph{Unique Correct Predictions Are Rare and Uncertain.}
M16 and M12 uniquely solve only 3.8--4.3\% of decisions, just 2--3 points above the noise floor. These wins are low-confidence (median confidence 0.59--0.61; full32 gold probability 0.23--0.33) and concentrated in Typed and AGNews (78\%), with none in J-Easy or J-Orig. Noise-floor disagreements show the same pattern: median confidence of 0.52--0.55, with 71--73\% from Typed and AGNews.

\paragraph{Smaller Models Also Lose Reliable Decisions.}
When full32 assigns the gold option probability $\geq0.8$, M24, M16, and M12 achieve 99.7\%, 84.1\%, and 70.4\% accuracy, respectively, while other full32 seeds reach 100\%. When full32 is confidently wrong ($p<0.2$), smaller models recover only 17--20\% of these decisions, compared with 0.7--0.9\% for other seeds. Their apparent gains therefore reflect greater disagreement with full32, rather than systematic correction of its errors. Overall, smaller models reproduce many of full32's correct decisions but lose substantial reliable capability without consistently correcting its mistakes.

\textbf{Takeaway.} Smaller decision models trained on their own do not learn a different solution; they learn a degraded version of the full model's. This suggests training them as \emph{parts} of the full model, sharing its weights and using its own output distribution as a target, so that each smaller model inherits as much of the full model's skill as its depth allows.

\section{Methodology}
\label{sec:method}

Section~\ref{sec:observation} shows that intermediate layers already encode sufficient information to identify the correct option, while smaller decision models are best viewed as prefixes of the full model that should learn to imitate it. Building on these findings, \method{} trains nested prefix sub-models within the full model using self-distillation to facilitate information sharing across depths (Sections~\ref{subsec:method_prefix} and~\ref{subsec:method_loss}). A lightweight gate then selects the exit depth for each decision based on question difficulty (Section~\ref{subsec:method_gate}).

\subsection{Prefix Sub-Models with a Shared Coda}
\label{subsec:method_prefix}

Let $f_0,\ldots,f_{31}$ be the layers of the decision model, and let $\mathbf{h}^{(e)}=f_e\circ\cdots\circ f_0(x)$ be the residual stream after layer $e$. We call the last two layers together with the decision head the \emph{coda} $C=\mathrm{head}\circ f_{31}\circ f_{30}$. An \emph{exit} after layer $e$ reads out the decision distribution
\begin{equation*}
p_e(\cdot\mid x)=\mathrm{softmax}\bigl(C(\mathbf{h}^{(e)})_{\mathcal{O}}/T_e\bigr),
\end{equation*}
where $\mathcal{O}$ is the set of option symbols and $T_e$ is a temperature fitted per exit. The sub-model $S_k$ with $k$ layers uses the exit after layer $e_k=k-3$, eg. $S_{12}$, $S_{16}$, and $S_{24}$ exit after layers 9, 13, and 21, and $S_{32}$ is the full model. This design follows Section~\ref{subsec:obs_where}. Prefixes keep the early feature-building layers that no smaller model can do without, and $S_{24}$ reaches into the decision band while dropping the late layers that carry no decision. The coda is shared so that every exit reads out through the same final layers and head. Because each sub-model is a prefix of the next, the states computed for a shallow exit are reused when a decision continues to a deeper exit, so stopping early never requires recomputation.

\subsection{Nested Training Objective}
\label{subsec:method_loss}

Following Section~\ref{subsec:obs_subset}, we train all sub-models jointly with the full model, so that they share its weights and learn from its output. Let $y$ be the gold option and $p_k$ the distribution of $S_k$. At each step we compute
\begin{align*}
\mathcal{L}={}&\mathrm{CE}(p_{32},y)+\sum_{k\in\mathcal{K}_t}\Bigl[(1-\alpha_t)\,\mathrm{CE}(p_k,y)\\
&\qquad+\alpha_t\,\KL\bigl(\mathrm{sg}(p_{32})\,\|\,p_k\bigr)\Bigr],
\end{align*}
where $\mathrm{sg}$ stops the gradient and the KL term is taken over the options of the decision. The full model is trained only on the gold option, and each sub-model learns both from the gold option and from the full model's distribution~\citep{hinton2015distilling}. This self-distillation term makes each sub-model imitate the full model, as Section~\ref{subsec:obs_subset} suggests. We linearly increase $\alpha_t$ from 0 to 0.5 over the first half of training, so that the full model's distribution becomes a target only once it is informative. The full model serves as its own teacher, so \method{} needs no external decision signal. All parameters are fine-tuned with the recipe of full32. 

\subsection{Choosing Depth per Decision}
\label{subsec:method_gate}

A single checkpoint now offers exits at intermediate depths, as well as the full model. To choose an exit for each decision, we attach a lightweight gate to each intermediate exit. The gate reads the hidden state $\mathbf{z}_e$ at the readout position immediately after the corresponding middle layer:
\begin{equation*}
g_e(x)=\sigma\bigl(\mathbf{w}_e^\top\mathbf{z}_e+b_e\bigr),
\end{equation*}
where $\sigma$ is the sigmoid function. Each gate is a logistic regression model trained to predict whether its corresponding exit answers correctly, using labels $c_e=\mathbf{1}[\arg\max p_e=y]$ and class-balanced weights. At inference, the model runs layer by layer. At each intermediate exit, it stops if $g_e(x)\geq\tau$, runs the coda, and returns $\arg\max p_e$; otherwise, it continues from the saved states. The final exit always answers.

Answering at exit $e$ costs $e+3$ layers: the prefix plus the coda. Because the gate reads the hidden state before the coda, evaluating it requires only a dot product and a sigmoid operation. A single threshold $\tau$ is shared across exits and selected on held-out data as the most aggressive value for which macro accuracy remains within $\delta$ points of always exiting at the final intermediate depth. 

\section{Experimental Results}
\label{sec:experiment}

We evaluate \method{} on three questions: (i) Does one nested checkpoint match models trained separately at each depth (Section~\ref{subsec:exp_nested})? (ii) Can the gate choose depth per decision without losing accuracy (Section~\ref{subsec:exp_gate})? (iii) Why does nested training help (Section~\ref{subsec:exp_why})?

\textbf{Setup.} All models start from Qwen3.5-4B and use the data, recipe, and benchmark of Section~\ref{sec:observation}. We compare \method{} with full32, M24, M16, and M12, each trained separately on its own. Every model, nested or separate, is trained with three seeds, which change the data order and, for \method{}, which sub-models are sampled. One \method{} run takes 80 minutes on a single H200, against 94 minutes for training the four separate models. At batch size one, the 12-, 16-, 24-, and 32-layer exits take 18.4, 23.8, 34.7, and 46.5\,ms per decision. For early exit, we count cost as the mean number of layers run per decision, including the coda. We fit each gate and its threshold on one benchmark-stratified half of the benchmark together with the calibration split, and report results on the other half. Each half is used once for testing. 

\subsection{One Checkpoint versus Separate Models}
\label{subsec:exp_nested}

\begin{table}[t]
\centering
\small
\setlength{\tabcolsep}{3.2pt}
\begin{tabular}{@{}rccr@{}}
\toprule
Layers & Separate & \method{} & $\Delta$ [95\% CI] \\
\midrule
32 & $\mathbf{84.3}_{\pm0.2}$ & $84.0_{\pm0.6}$ & $-0.3$ [$-1.3$, $+0.8$] \\
24 & $83.5_{\pm0.8}$ & $\mathbf{83.7}_{\pm0.3}$ & $+0.2$ [$-1.1$, $+1.5$] \\
16 & $72.8_{\pm0.9}$ & $\mathbf{74.7}_{\pm0.9}$ & $+1.8$ [$-0.1$, $+3.8$] \\
12 & $61.5_{\pm1.5}$ & $\mathbf{64.4}_{\pm0.4}$ & $\mathbf{+2.9}$ [$+0.7$, $+5.3$] \\
\bottomrule
\end{tabular}
\caption{Benchmark average (\%) of separately trained models and of the exits of one \method{} checkpoint (mean$_{\pm\text{sd}}$ over 3 seeds). $\Delta$: \method{} minus separate, with a CI that resamples decisions and seeds.}
\label{tab:main}
\end{table}

\paragraph{One Checkpoint Matches or Outperforms Separate Models at Every Depth.}
Table~\ref{tab:main} compares the four exits of \method{} with independently trained models at matching depths. No exit performs significantly worse, while the shallower exits achieve higher accuracy. The largest gain occurs at 12 layers (64.4\% vs.\ 61.5\%, +2.9 points), with a confidence interval excluding zero even when resampling seeds. At 16 layers, the gain is 1.8 points but does not reach significance. The 24-layer exit achieves 83.7\%, comparable to M24 (+0.2 points) and full32 ($-0.6$ points), while using 25\% fewer layers and reducing latency from 46.5 to 34.7\,ms. At 32 layers, accuracy is only 0.3 points below full32, indicating that joint training imposes little cost on the full model. Thus, a single 80-minute training run replaces four separate runs while yielding the best accuracy at 12, 16, and 24 layers.

\subsection{Choosing Depth per Decision}
\label{subsec:exp_gate}

\begin{table}[t]
\centering
\small
\setlength{\tabcolsep}{3.2pt}
\begin{tabular}{@{}lrrrrrr@{}}
\toprule
 & & & & \multicolumn{3}{c}{Mean layers} \\
\cmidrule(l){5-7}
Rule & $\delta$ & $\Delta$Avg & $\Delta$L4 & All & Easy & Hard \\
\midrule
Fixed exit 13 & -- & $-9.89$ & $-4.49$ & 16.0 & 16.0 & 16.0 \\
Fixed exit 21 & -- & 0.00 & 0.00 & 24.0 & 24.0 & 24.0 \\
\midrule
Max-prob & 0.25 & $-0.30$ & $-0.04$ & 23.8 & 23.4 & 27.8 \\
Confidence gate & 0.25 & $-0.39$ & $-0.06$ & 23.2 & 22.3 & 27.5 \\
State + confidence & 0.25 & $-0.26$ & $+0.11$ & 23.4 & 19.7 & 27.9 \\
\textbf{State gate (ours)} & 0.25 & $-1.56$ & $+0.08$ & \textbf{21.6} & 20.2 & \textbf{24.0} \\
\midrule
Max-prob & 2 & $-1.35$ & $-0.12$ & 21.3 & 19.4 & 26.9 \\
Confidence gate & 2 & $-1.36$ & $-0.13$ & 21.1 & 19.0 & 26.6 \\
State + confidence & 2 & $-1.37$ & $-0.09$ & 20.2 & \textbf{15.4} & 27.3 \\
\textbf{State gate (ours)} & 2 & $-1.68$ & $+0.03$ & \textbf{20.2} & 17.5 & \textbf{23.9} \\
\bottomrule
\end{tabular}
\caption{Early exit over exits 9, 13, and 21 of \method{} (seed 42). $\Delta$Avg: change in benchmark average relative to always exiting after layer 21 (83.8\%). $\Delta$L4: accuracy change on the four largest benchmarks (AGNews, WildJailbreak, Typed, J-Hard; 96\% of decisions), weighted by decisions. Mean layers include the coda and every readout computed. Easy: AGNews and WildJailbreak; hard: Typed and J-Hard. Max-prob stops when the calibrated top probability clears the threshold; the confidence gate is a logistic regression on top probability, margin, and entropy.}
\label{tab:gate}
\end{table}

\paragraph{The Hidden State Predicts Whether an Exit Is Correct.} We first measure how well each signal separates correct from wrong answers at an exit, using AUROC on the test halves. The state gate reaches 0.86, 0.89, and 0.88 at exits 9, 13, and 21, against 0.85, 0.86, and 0.84 for the calibrated top probability, and 0.86, 0.86, and 0.84 for the confidence gate that combines top probability, margin, and entropy. The hidden state thus carries more information about correctness than the exit's own confidence, especially at exit 13 and 21, and it is available before the coda is computed.

\paragraph{Stopping Rules.} We compare the state gate with three rules based on the exit distribution $p_e$: \emph{Max-prob}, which uses the calibrated top probability; \emph{confidence gate}, a logistic regression on the top probability, top-two margin, and entropy; and \emph{State + confidence}, which combines these features with the hidden state $\mathbf{z}_e$. All rules use the same training labels, class-balanced weights, and fitting data. They evaluate exits 9, 13, and 21 in order, stopping at the first score above $\tau$ or answering at exit 21 otherwise; $\tau$ is selected as in Section~\ref{subsec:method_gate}. The key difference is inference cost: distribution-based rules must run the coda to obtain $p_e$, adding two layers at each checked exit, whereas the state gate reads $\mathbf{z}_e$ before the coda and runs it only once, at the selected exit. Fixed exits 13 and 21 serve as references.

\paragraph{The State Gate Gives the Best Trade-off.} Table~\ref{tab:gate} compares the gating rules at two accuracy thresholds. The state gate achieves the lowest average depth at both $\delta=0.25$ and $\delta=2$, using 21.6 and 20.2 layers, respectively, compared with 24 layers for the fixed exit. On hard benchmarks, it uses 23.9--24.0 layers, whereas confidence-based rules incur two additional layers per readout, resulting in an average depth of 26.6--27.9 layers. On the four largest benchmarks, which account for 96\% of decisions, the state gate preserves the fixed-exit accuracy, with changes of only $+0.08$ and $+0.03$ points at the two thresholds. The other rules also remain within 0.13 points of the fixed exit, but achieve savings differently: confidence-based rules save computation on easy decisions while spending more on hard ones, whereas the state gate reduces depth on easy decisions (to 20.2 and 17.5 layers) without incurring additional computation on hard decisions. Randomly mixing exits 13 and 21 to match the same average depth instead reduces accuracy by 3.0 and 4.7 points.

\paragraph{The Accuracy Drop Is Concentrated on Small Benchmarks.}
The state gate's macro-accuracy drop ($-1.56$ and $-1.68$ points) comes almost entirely from the three smallest benchmarks, which contain only 430 of 12,351 decisions but account for three of seven macro-averaged benchmarks. At $\delta=0.25$, it misses seven additional decisions on these benchmarks but loses none on larger ones. Compute savings depend on traffic mix: average depth ranges from 24 layers on hard-only traffic to 20.2 on easy-only traffic (22.1 for an even mix), and to 17.5 layers at $\delta=2$ (20.7 for an even mix; Figure~\ref{fig:teaser}, right). Since the selected threshold varies across training splits (0.80 vs.\ 0.92), it should be calibrated on held-out deployment-like traffic. Adding the full model as a final exit further improves the trade-off: at $\delta=2$, the gate matches its accuracy on the four largest benchmarks ($-0.01$ points) while using 23.9 layers on average, 19.8 on easy decisions, and 31.3 on hard ones.

\subsection{Why Nested Training Helps}
\label{subsec:exp_why}

\begin{table}[t]
\centering
\small
\setlength{\tabcolsep}{3.4pt}
\begin{tabular}{@{}lrrrrrr@{}}
\toprule
 & & \multicolumn{2}{c}{Only correct} & & \multicolumn{2}{c}{Acc $\mid p_{\text{full}}$} \\
\cmidrule(lr){3-4}\cmidrule(l){6-7}
Model & Acc & full & small & Rec. & $\geq$.8 & $<$.2 \\
\midrule
full32, seeds 43/44 & 84.3 & 1.6 & 1.7 & .975 & 100.0 & 0.8 \\
\midrule
M12 & 61.5 & 26.5 & 3.8 & .657 & 69.5 & 15.8 \\
\method{} 12 & 64.4 & \textbf{23.4} & 3.6 & \textbf{.688} & \textbf{72.8} & 14.8 \\
\midrule
M16 & 72.8 & 15.1 & 3.6 & .785 & 83.3 & 17.3 \\
\method{} 16 & 74.7 & \textbf{13.2} & 3.5 & \textbf{.810} & \textbf{86.0} & 17.9 \\
\midrule
M24 & 83.5 & 2.9 & 2.2 & .956 & 99.4 & 3.8 \\
\method{} 24 & 83.7 & 3.4 & 2.8 & .950 & 98.8 & 3.4 \\
\method{} 32 & 84.0 & 2.7 & 2.4 & .961 & 99.3 & 2.8 \\
\bottomrule
\end{tabular}
\caption{Decision-by-decision comparison with full32 (seed 42), as in Table~\ref{tab:overlap}, averaged over 3 seeds of each model. \method{} $k$ is the $k$-layer exit.}
\label{tab:overlap_nested}
\end{table}

\paragraph{Shallow Exits Keep More of the Full Model's Skill.} Section~\ref{subsec:obs_subset} found that separately trained small models are degraded copies of the full model. Table~\ref{tab:overlap_nested} shows that nesting reduces this degradation rather than adding new skills. At 12 layers, the exit wins no more decisions of its own than M12 does (3.6\% vs.\ 3.8\%). Instead, it loses fewer of full32's correct decisions (23.4\% vs.\ 26.5\%), so its recall of full32's skill rises from 0.66 to 0.69. On decisions full32 is confident about, it is right 72.8\% of the time, against 69.5\% for M12. The 16-layer exit shows the same pattern: recall rises from 0.79 to 0.81, and accuracy on confident decisions from 83.3\% to 86.0\%, with no more wins of its own (3.5\% vs.\ 3.6\%). This is the effect the self-distillation term targets: the shallow exit imitates the full model and keeps more of what the full model knows.

\paragraph{Deep Exits Stay Close to the Full Model.} At 24 and 32 layers, the nested exits keep as much of full32's skill as M24 (recall 0.95--0.96 vs.\ 0.96), and their own wins (2.4--2.8\%) are close to those of M24 (2.2\%). Nesting neither adds nor removes skill at these depths. This matches Table~\ref{tab:main}, where the deep exits tie their separate counterparts. The benefit of nesting at depth is therefore cost, not accuracy: one checkpoint provides the 24-layer exit and the full model together, and the gate can choose between depths for each decision.

\begin{figure}[t]
\centering
\includegraphics[width=\linewidth]{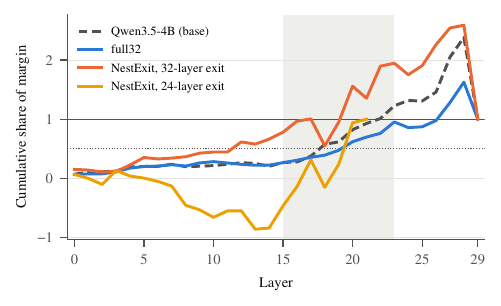}
\caption{Cumulative share of the decision margin contributed by layers up to $\ell$, measured with integrated gradients on 348 decisions. The 24-layer exit of \method{} reads out after layer 21. The shaded band marks layers 15--23.}
\label{fig:nested_where}
\end{figure}

\paragraph{Where Does \method{} Write the Decision?}
Figure~\ref{fig:nested_where} shows that nested training shifts decision formation toward earlier layers, reversing the delay caused by plain fine-tuning. In the 32-layer exit, layers 0--16 account for half of the final margin (bootstrap range 15--16), and layers 0--20 account for 80\% (19--20), compared with layers 20 and 23 in full32. The first 20 layers contribute 77\% of the margin, versus 47\% in full32, so the full model largely commits to its decision within the prefix shared by the 24-layer exit. Later layers mainly rescale this decision: their cumulative contribution peaks at 2.5$\times$ the final margin before layer 29 removes the excess. Jointly ablating layers 24--29 increases the margin by 4\% and reduces accuracy by only 0.6 points. The 24-layer exit follows a similar pattern: layers 15--21 form its decision, with layer 20 alone contributing 48\%. Thus, nested training concentrates decision formation in layers shared across deeper exits.

\section{Conclusion}
\label{sec:conclusion}

We studied adaptive computation for System One decision-making in LLMs. Our analysis reveals that decisions primarily emerge in middle layers, with later layers contributing little additional information, and that smaller independently trained models largely replicate the full model's decisions. Motivated by these findings, \method{} jointly trains nested prefix sub-models via self-distillation and employs a hidden-state gate for adaptive early exiting. Across three seeds, a single checkpoint matches or outperforms independently trained models at all exit depths, improving 12-layer accuracy by 2.9 points and approaching the 32-layer model within 0.6 points at 24 layers. Adaptive exiting further reduces computation by 16--27\% on easy decisions without sacrificing accuracy, while retaining fixed-depth exits for harder cases. These results highlight the benefits of nested training for aligning shallow exits with full-model decisions. Future work will investigate end-to-end latency, gate calibration under real-world traffic, and scaling to larger models and finer-grained exits.



\bibliography{custom}

\end{document}